\RequirePackage{fix-cm}
\documentclass[twocolumn]{svjour3}          % twocolumn

\smartqed  % flush right qed marks, e.g. at end of proof

\usepackage{amsmath}
\usepackage{amssymb}
\usepackage{array}
\usepackage{booktabs} % For formal tables
\usepackage{cite}
\usepackage{flushend}

\usepackage{textcomp} % to avoid \micro and \perthousnad warning by the gensymb package
\usepackage{gensymb}

\usepackage{graphicx}
\usepackage{mathtools, cuted}
\usepackage{microtype}
\usepackage{multirow}
\usepackage{rotating}
\usepackage{tablefootnote}

\usepackage{url}

\begin{document}
%\sloppy

\title{Companion Unmanned Aerial Vehicles: A Survey}

\author{Chun Fui Liew and Takehisa Yairi}

\newcommand{\etal}{\mbox{\emph{et al.}}}

\newcommand{\nparagraph}[1]{\vspace{2mm}\noindent\textbf{#1:}}

\institute{Chun Fui Liew \at
               The University of Tokyo, Japan \\
               \email{liew@g.ecc.u-tokyo.ac.jp }
           \and
           Takehisa Yairi \at
              The University of Tokyo, Japan \\
              \email{yairi@g.ecc.u-tokyo.ac.jp}
}

\date{}

\maketitle

\begin{abstract}
Recent technological advancements in small-scale unmanned aerial vehicles (UAVs) have led to the development of companion UAVs. Similar to conventional companion robots, companion UAVs have the potential to assist us in our daily lives and to help alleviating social loneliness issue. In contrast to ground companion robots, companion UAVs have the capability to fly and possess unique interaction characteristics. Our goals in this work are to have a bird's-eye view of the companion UAV works and to identify lessons learned and guidelines for the design of companion UAVs. We tackle two major challenges towards these goals, where we first use a coordinated way to gather top-quality human-drone interaction (HDI) papers from three sources, and then propose to use a perceptual map of UAVs to summarize current research efforts in HDI. While being simple, the proposed perceptual map can cover the efforts have been made to realize companion UAVs in a comprehensive manner and lead our discussion coherently. We also discuss patterns we noticed in the literature and some lessons learned throughout the review. In addition, we recommend several areas that are worth exploring and suggest a few guidelines to enhance HDI researches with companion UAVs.
\end{abstract}


% ========== ========== ==========
\section{Introduction}
\label{sec:introduction}
% ========== ========== ==========

Companion robots are robots designed to have social interaction and emotional connection with people. For examples, companion robots like Paro (a therapeutic robotic seal for elderly)~\cite{social-robot_wada_tro07}, EmotiRob (a companion robotic bear for children)~\cite{social-robot_tallec_sbh11}, Aibo (a companion robotic dog)~\cite{social-robot_csaba_jais18}, and Jibo (a companion robot in home)~\cite{social-robot_jibo_times17} could interact socially with people. One potential application of companion robots is that they could be our personal assistance. One could also argue that companion robots are similar to pets and they help to alleviate social isolation or loneliness issues.

Recently, technological advancements have led to a new class of companion robots---a companion unmanned aerial vehicle (UAV). Compared to conventional companion robots, companion UAVs have some distinctive characteristics---notably their capability to fly (hence their new interaction capability) and their strict design constraints such as safety concern, noise issue, flight time and payload limitation. While there are still many technical difficulties and social design questions, various concepts, design sketches, and prototypes have been proposed to demonstrate the idea of a companion UAV, including a flying jogging companion~\cite{uav_hri-platform_graether_12}, a flying assistant that could interact with us in daily tasks~\cite{social-interaction_vink_youtube14},  a human-centered designed drone aiming to fly in a human crowd environment~\cite{social-interaction_yeh_hai17}, a flying smart agent that could assist users through active physical participation~\cite{social-interaction_agrawal_ubicomp15}, flying ``fairies''~\cite{companion_uav-murphy-ar11, companion_uav-duncan-hri10} and flying lampshades~\cite{uav_hri-platform_flying-lampshade-1, uav_hri-platform_flying-lampshade-2} that could dance with human on the stage, a flying ball for augmented sports~\cite{uav_hri-platform_nitta_14}, a flying humanoid balloon to accompany children~\cite{uav_hri-platform_cooney_12}, a companion UAV that can react to human emotions~\cite{companion_uav-malliaraki-vimeo17}, and a moving projector platform for street games~\cite{social-interaction_kljun_chiplay15}.

Our goals in this survey are to have a bird's-eye view of the companion UAV works and to identify lessons learned, guidelines, and best practices for the design of companion UAVs. There are two major challenges towards these goals: (i) to find a coordinated way to identify top-quality HDI works from a huge amount of available literature, and (ii) to find a suitable framework or organizing principle to discuss the vast aspects of existing human-drone interaction (HDI) works. 

To tackle the first challenge, i.e., to cover top-quality companion UAV works as comprehensive as possible in this review, we use a coordinated way to gather HDI papers from three major sources. First, we systematically identified 1,973 high-quality UAV papers from more than forty-eight thousand general papers that have appeared in the top robotic journals and conferences since 2001. In a nutshell, this identification process involves a few steps consist of automated and manual processes (more details in Section~\ref{sec:uav_paper_identification_process}).
%In a nutshell, we first used a script to automatically collect more than forty-eight thousand instances of title and abstract from the official journal/conference web pages. Then, we designed a list of keywords to search UAV papers systematically from the collected titles and abstracts. Finally, we performed a manual screening to reject some non-UAV papers. Specifically, We read the abstract, section titles, related works, and experiment results of all papers, and if a paper passes preset criteria, we consider it a UAV paper.\footnote{Details on the paper category analysis can be found in our previous survey paper~\cite{uav-survey_liew_arxiv_17}. Details about the online sharing and regular updates can be found in Appendix~\ref{appendix:uav_database_update_and_online_sharing}.} 
Second, from the identified UAV papers, we tagged the papers with several key topics, studied those related to the topic of HRI and analyzed their references, and continued to track down HDI-related papers. Third, we included HDI papers recommended by reviewers during our past journal submission.

To tackle the second challenge, i.e., to find a suitable framework to discuss the vast aspects of existing HDI works, we propose to use a perceptual map of UAVs (more details in Section~\ref{sec:perceptual_map_of_uavs}) as a high-level framework to organize current research efforts in HDI. In the proposed perceptual map, we categorize UAVs based on the degree of autonomy and the degree of sociability. This basic categorization leads to four distinct categories, namely remotely-controlled UAV, autonomous UAV, social UAV, and companion UAV. Looking at the research and development of companion UAVs with this perceptual map, we can find two main direction of on-going efforts. Moreover, we find this perceptual map easy to understand and lead our discussion coherently. 

This work emphasizes on the proximate interaction between a human and a companion UAV in the HDI field. Note that proximate interaction is also called collocated interaction in some literature~\cite{social-interaction_wojciechowska_hri19, social-interaction_hedayati_hri18}. In the following sections, we first briefly explain the definition of a UAV and different type of UAVs (Section~\ref{sec:uav_background}) in order to facilitate the discussion in this work. Next, we describe methodology we used to identify top-quality UAV papers from the literature (Section~\ref{sec:uav_paper_identification_process}). Then, we discuss the perceptual map of UAVs (Section~\ref{sec:perceptual_map_of_uavs}), followed by discussion on research efforts in realizing companion UAVs from the engineering (Section~\ref{sec:from_remote_controlled_to_autonomous_uavs}) and sociability (Section~\ref{sec:from_remote_controlled_to_social_uavs}) perspectives. In Section~\ref{sec:observation_and_lessons_learned} and Section~\ref{sec:guidelines_and_recommendations}, we discussion several observation and lessons learned, along with guidelines and recommendations for realizing a companion UAV. Section~\ref{sec:conclusion} draw conclusion about future research directions for companion UAVs.

% ========== ========== ==========
\section{UAV Background}
\label{sec:uav_background}
% ========== ========== ==========

We first explain the UAV definition and introduce some common UAV types to facilitate the following discussion. We recommend the handbook of UAVs~\cite{handbook_of_uav} if readers are interested in the more technical details of UAV.

% what UAV means?
\subsection{UAV Definition}
\label{subsec:uav_definition}

UAVs, commonly known as drones, are aircraft that can perform flight missions without a human pilot onboard~\cite{faa_uav}. In general, UAVs can be viewed as flying robots. The UAV's degree of autonomy varies but often modern UAVs are able to hover stably at a point in 3D space. UAVs with a higher degree of autonomy offer more functions like automatic take-off and landing, path planning, and obstacle avoidance. In the literature, UAVs have several other names such as micro aerial vehicle (MAV), unmanned aerial system (UAS), vertical take-off and landing aircraft (VTOL), multicopter, rotorcraft, and aerial robot. In this work, we will use the ``UAV'' and ``drone'' terms interchangeably.

% what can be classified as a UAV?
\subsection{UAV Types}
\label{subsec:uav_types}

Conventionally, UAVs can be classified into fixed-wing, multirotor, blimp, or balloon types based on their flying principle. In the end of this review, one could observe that most UAV prototypes we described in this work are multirotor UAVs. We speculate this is due to the availability of multirotor type UAVs in the market. In Section~\ref{sec:observation_and_lessons_learned}, we will have a more rigorous discussion arguing that the blimp or balloon UAVs could be a better form for companion UAVs.

\begin{figure}[tb]
  \centering
  \includegraphics[trim={9cm 10cm 8cm 8cm},clip,width=0.42\textwidth]{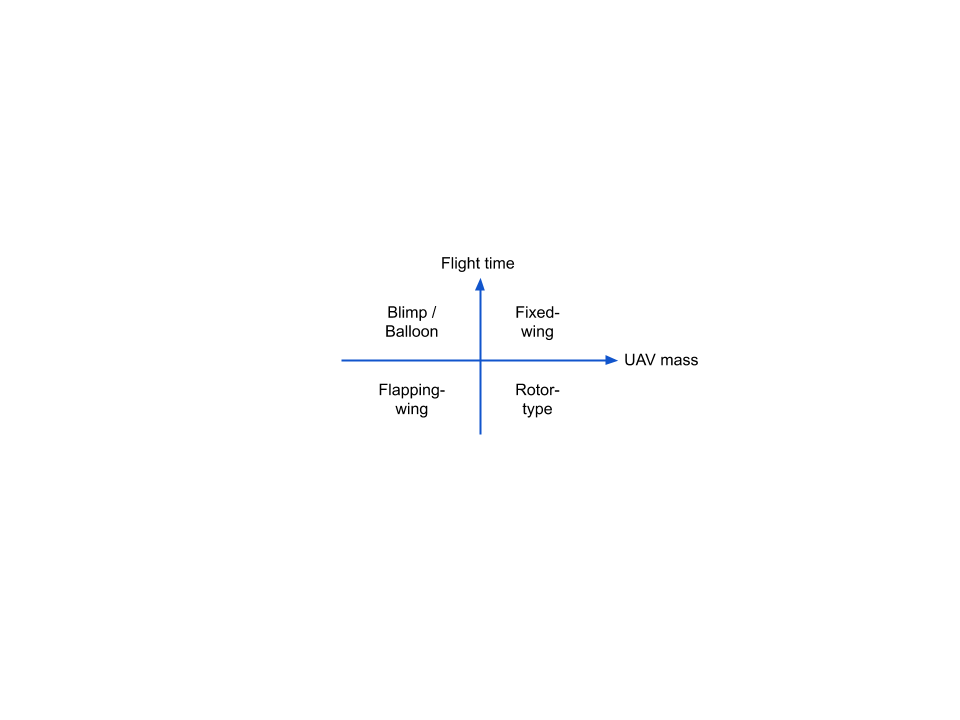}
  \caption{UAV types based on flight time and UAV mass (inspired by Floreano \& Wood~\cite{uav_suvey_floreano_15}).}
  \label{fig:uav_categories}
\end{figure}

It is worth noting that Floreano \& Wood~\cite{uav_suvey_floreano_15} have classified UAVs based on the flight time and UAV mass (a simplified plot is shown in Fig.~\ref{fig:uav_categories}). In general, flapping-wing UAVs are small and have a short flight time. Blimp/balloon UAVs are lightweight and have a longer flight time. Rotor-type and fixed-wing UAVs are usually heavier. In Section~\ref{sec:observation_and_lessons_learned}, we will have more discussion about the safety and noise issues of different types of UAVs.

\section{UAV Paper Identification Process}
\label{sec:uav_paper_identification_process}
% ========== ========== ==========

The UAV papers identification process involves three major steps. We first used a script to automatically collect more than forty-eight thousand instances of title and abstract from fourteen top journal/conference web pages since 2001 (the seven journals include IEEE Transactions on Robotics (TRO), IEEE/ASME Transactions on Mechatronics (TME), The International Journal of Robotics Research (IJRR), IAS Robotics and Autonomous Systems (RAS), IEEE Robotics and Automation Letters (RA-L), ACM Journal on Interactive, Mobile, Wearable and Ubiquitous Technologies (IMWUT), and ACM Transactions on Human-Robot Interaction (THRI); the seven conferences include IEEE International Conference on Intelligent Robots and Systems (IROS), IEEE International Conference on Robotics and Automation (ICRA), ACM/IEEE International Conference on Human-Robot Interaction (HRI), IEEE International Workshop on Robot and Human Communication (ROMAN), ACM Conference on Human Factors in Computing Systems (CHI), ACM International Conference on Ubiquitous Computing (UbiComp), and ACM International Conference on Human-Computer Interaction with Mobile Devices and Services (MobileHCI). We also manually reviewed the hard copies of the IROS and ICRA conferences' table of contents from 2001 to 2004, as we find that not all UAV papers in those years are listed on the website (IEEE Xplore). 

Then, we designed a list of keywords (Table~\ref{tab:drone-keywords}) to search drone papers systematically from the titles and abstracts collected in the first step. Note that we searched for both the full name of each keyword (e.g., Unmanned Aerial Vehicle) and its abbreviation (i.e., UAV) with an automatic program script. The keywords include most of the words that describe a UAV. For example, the word ``sUAV'' (small UAV) could be detected by the keyword ``UAV''. Similarly, the word ``quadcopter'' or ``quadrotor'' could be detected by the keyword ``copter'' or ``rotor''. As long as one of the keywords is detected, the paper will pass this automated screening process.

\begin{table}[tbp]
  \centering
  \caption{35 keywords used to search drone papers systematically from the collected titles and abstracts.}
  \label{tab:drone-keywords}
  \scriptsize
  \begin{tabular}{|l|l|l|l|}
    \hline
    acrobatic & bat       & flight   & rotor                                \\
    aerial    & bee       & fly      & rotorcraft                           \\
    aero      & bird      & flying   & soar                                 \\
    aeroplane & blimp     & glide    & soaring                              \\ \cline{4-4}
    air       & copter    & glider   & micro aerial vehicle                 \\
    aircraft  & dragonfly & gliding  & unmanned aerial vehicle              \\
    airplane  & drone     & hover    & unmanned aircraft system             \\
    airship   & flap      & hovering & vertical takeoff and landing         \\ \cline{4-4}
    balloon   & flapping  & kite     & MAV, UAV, UAS, VTOL                  \\
    \hline
  \end{tabular}
\end{table}

Finally, we performed a manual screening to reject some non-drone papers. We read the abstract, section titles, related works, and experiment results of all the papers from the second step. If a paper passes all the five criteria below, we consider it a drone paper for this survey.
%In a nutshell, we first used a script to automatically collect more than forty-eight thousand instances of title and abstract from the official journal/conference web pages. Then, we designed a list of keywords to search UAV papers systematically from the collected titles and abstracts. Finally, we performed a manual screening to reject some non-UAV papers. Specifically, We read the abstract, section titles, related works, and experiment results of all papers, and if a paper passes preset criteria, we consider it a UAV paper.
\footnote{Details on the paper category analysis can be found in our previous survey paper~\cite{uav-survey_liew_arxiv_17}. Details about the online sharing and regular updates can be found in Appendix~\ref{appendix:uav_database_update_and_online_sharing}.} 

\begin{enumerate}
  \item The paper must have more than two pages; we do not consider workshop and poster papers.
  \item The paper must have at least one page of flight-related results. These can be either simulation / experiment results, prototyping / fabrication results, or insights / discussion / lesson learned. One exception is a survey/review paper, which normally does not present experiment results. Papers with details or photos of the UAV hardware are a plus. Note that the experiment results do not necessarily need to be a successful flight, e.g., flapping wing UAVs normally have on-the-bench test results.
  \item In topics related to computer vision, the images must be collected from a UAV's onboard camera rather than a manually moving camera.
  \item In topics related to computer vision, the images must be collected by the authors themselves. This is important, as authors who collected the dataset themselves often provide insights about their data collection and experiment results.
  \item The paper which proposes a general method, e.g., path planning, must have related works and experiment results on drones. This is important, as some authors mention that their method can be applied to a UAV, but provide no experiment result to verify their statement.
\end{enumerate}

It is interesting to note that using the keyword ``air'' in the second step increases the number of false entries (since the keyword is used in many contexts) but helps to identify some rare drone-related papers that have only the keyword ``air'' in the title and abstract. By manually filtering the list in the third step, we successfully identified two of these drone papers~\cite{uav-background_quadcopter_latscha_iros14, uav-paper_butzke_iros_15}. Similarly, using the keyword ``bee'' can help to identify a rare drone paper~\cite{uav-paper_das_ras_16}. On the other hand, we chose not to use the keyword of ``wing'' because it causes many false entries like the case of ``following'', ``knowing'', etc.

\section{Perceptual Map of UAVs}
\label{sec:perceptual_map_of_uavs}

Having a framework that could cover all the related companion UAV works in both engineering and social interaction topics is challenging, as these papers have different motivation, methodology, and results. Companion UAV works focusing on engineering usually emphasize on devising new hardware designs or new autonomous functions, while companion UAV works focusing on social interaction studies usually emphasize on participatory design and social experiments with users. To this end, we propose to categorize related works in this survey based on a perceptual map with the degree of autonomy (corresponding to engineering works) and the degree of sociability (corresponding to social interaction works) (Fig.~\ref{fig:uav_categories})

\subsection{The Four UAV Categories}
\label{subsec:the_four_uav_categories}

\begin{figure*}[tb]
  \centering
  \includegraphics[trim={0cm 8cm 0cm 7cm},clip,width=0.75\textwidth]{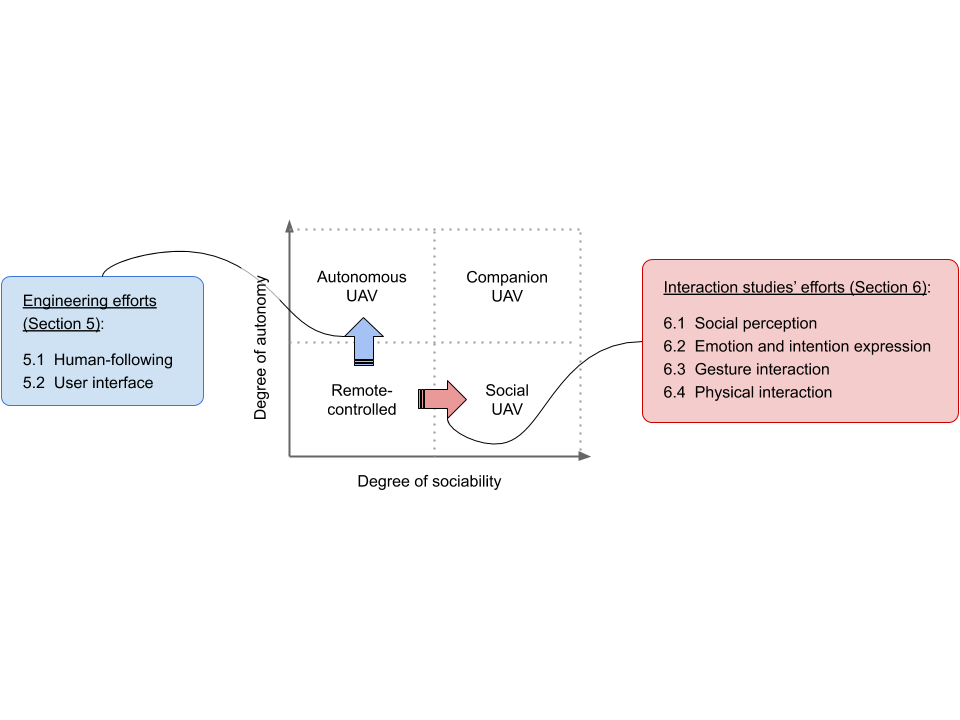}
  \caption{Perceptual map for UAVs based on the degree of autonomy and sociability, with the major topics found in the literature.}
  \label{fig:uav_perceptual_map}
\end{figure*}

The perceptual map of UAVs has four categories: remote-controlled UAV, autonomous UAV, social UAV, and companion UAV. Traditionally, UAVs are controlled manually by human operators and have low degrees of autonomy and sociability. Gradually, along the vertical axis of degree of autonomy, researchers have been improving the autonomy aspects of UAVs, such as better reactive control with more sensors and better path planning algorithms. Essentially, autonomous UAVs are less dependent on human operators and are able to perform simple flight tasks autonomously.

At the same time, along the horizontal axis of degree of sociability, researchers have been improving the social aspects of UAVs, such as designing UAV movements that are more comfortable for humans or building intuitive interfaces for us to understand UAVs' attention better. Most HRI researchers focus on the social aspects of UAVs and usually perform user studies using Wizard of Oz\footnote{A common experiment setting used by researchers, where participants interact with a robot that participants believe to be autonomous, but in fact it is being manually controlled by a human behind the scene.} experiments. Different from autonomous UAVs, in which its main purpose is to achieve a task efficiently from an engineering point of view, social UAV aims to work with human harmonically, i.e., ease user acceptance and relieve user's cognitive burden. For example, a ``social'' fire-fighting drone might need to have a design that make nearby human understanding its purpose of fire-fighting during emergency~\cite{social-interaction_khan_chi19}.

We first coined the phrase ``companion UAV'' and consider a companion UAV as one that possesses high degrees of both autonomy and sociability~\cite{liew_dissertation_16}. In addition to the autonomy skills, such as motion planning and obstacle avoidance, companion UAVs must also feature sociability skills such as making users feel safe and understand their intention. It is worth noting that the term ``companion UAV'' used in a prior work~\cite{companion_uav-hrabia-ros17} has a different meaning, where the UAV was designed to support a ground robot but not to interact with a person.

To consolidate the idea of perceptual map, we can use a package-delivery drone as an example. In the autonomous UAV sense, the package-delivery drone focuses on accomplishing the delivery task from the engineering perspective. Note that this alone is a challenging task as one needs to figure out how to perform the flight efficiently, how to detect the landing location, how to avoid obstacles during the flight, etc. On the other hand, in the sociable UAV sense, the package-delivery drone should also acknowledge people for successful interactions with people, i.e., signaling to the person that it has seen them at a certain distance and a range of time. This design aspect has also been raised and investigated recently by Jensen~\etal~\cite{social-interaction_jensen_chi18}.

\subsection{Link with Previous Study}
\label{subsec:link_with_previous_study}

It should be noted that ``social UAV'' has been mentioned in the past literature frequently. One representative example is an survey paper of social UAV~\cite{social-interaction_baytas_chi19}. In this study, Bayta{\c s}~\etal~define ``social drone'' as autonomous flyers operate near to human users. Literally, this definition is similar to the definition of ``companion UAV'' here, but upon careful investigation, their true meaning of ``social drone'' is closer to the meaning of ``social UAV'' mentioned in the perceptual map here. Most papers considered by Bayta{\c s}~\etal~were not truly autonomous, i.e., a drone with pre-programmed behavior and motions is considered as autonomous by them. In our opinion, their categorization is not precise enough, e.g., while the teleoperated drone~\cite{social-interaction_jones_dis16} is categorized as a ``social drone'' by them, we consider that teleoperated drone a remotely-controlled drone in our context here as the drone is neither autonomous nor social. They also included poster and short papers in their review, but it is unclear how they categorize some short and poster papers that lack of implementation details. In contrast, our work here cover more papers in a more coordinated and systematic way.

\subsection{Research Efforts Towards Companion UAVs}
\label{subsec:research_efforts_towrads_companion_uavs}

Designing companion UAVs is challenging as it involves both technical/engineering issues and social/emotional design questions. We believe that because of this reason, most UAV works identified in this survey focus on a single issue, either an engineering or social issue, rather than having an ambitious goal to tackle both issues in a paper. As shown in Fig.~\ref{fig:uav_perceptual_map}, there are two main efforts for realizing companion UAVs in the literature, where the first one moves from the remote-controlled UAV to the autonomous UAV direction (blue arrow), and the second one moves from the remote-controlled UAV to the social UAV direction (red arrow).

The blue arrow in Fig.~\ref{fig:uav_perceptual_map} signifies efforts of robotic developers in realizing companion UAVs. In the literature, these efforts have a distinctive feature where the authors include engineering details, be it about the UAV hardware, control algorithms, or visual tracking methods. From the identified companion UAV works in Section~\ref{sec:uav_paper_identification_process}, the topics of human-following UAVs and user interface clearly emerge in this area.
%Human-following is an important skill for companion UAVs. Without the capability to follow a human, companion UAVs might not be considered a ``companion UAV" after all. Designing natural and intuitive control interfaces for companion UAVs is also important for HDI.
In Section~\ref{sec:from_remote_controlled_to_autonomous_uavs}, we will discuss these sub-topics in more details.

The red arrow in Fig.~\ref{fig:uav_perceptual_map} signifies efforts of HDI researchers in realizing companion UAVs. In the literature, these efforts have a distinctive feature where the authors performed Wizard-of-Oz experiments or carried out online surveys by using HDI videos. From the identified companion UAV works in Section~\ref{sec:uav_paper_identification_process}, the topics of social perception of UAVs, emotion and intention expression of UAVs, gesture and physical interaction with UAVs clearly emerge in this area.
%Emotion expression or intention expression of UAVs through light, sound, or motion is crucial for human to understand UAVs in interactive or collaborative tasks. Social interaction studies are useful to determine the appropriate social norm parameters, such as flying height, proximate distance, UAV shapes, etc. so that UAVs can work with human harmonically.
In Section~\ref{sec:from_remote_controlled_to_social_uavs}, we will discuss these sub-topics in more details.

% ========== ========== ==========
\section{From Remote-Controlled to Autonomous UAVs}
\label{sec:from_remote_controlled_to_autonomous_uavs}
% ========== ========== ==========

Developing companion UAVs that are autonomous and sociable is not a straightforward task. Most companion UAV works focus on one topic for realizing a companion UAV. In this section, we aim to summarize engineering efforts for realizing a companion UAV, with focus on the human-following and control interface topics.

\subsection{Human Following UAVs}
\label{subsec:human_following_uavs}

Pestana~\etal~used a UAV's onboard camera and object tracking algorithm to realize a human following application~\cite{uav_hri-human-tracking_pestana_13}. Higuchi~\etal~also performed human following with the UAV's onboard camera by using a color-based particle filter~\cite{uav_hri-human-tracking_higuchi_11}. On the other hand, Papachristos~\etal~demonstrated a human tracking application with the UAV's onboard stereo camera~\cite{uav-background_quadcopter_papachristos_iros15}. All the proposed prototypes focused on the functional designs of the system and did not carry out social interaction experiments. Moreover, these systems have special requirements such as manual initialization of the user location~\cite{uav_hri-human-tracking_pestana_13}, the necessity for the user to wear a shirt of a specific color~\cite{uav_hri-human-tracking_higuchi_11}, or the necessity for the user to move (so that the image tracker starts working)~\cite{uav-background_quadcopter_papachristos_iros15}.

By integrating a visual SLAM technique and a vision-based human tracking algorithm, Lim \& Sinha presented a UAV that can map the human walking path in real time~\cite{uav_hri-human-tracking_lim_15}. On the other hand, Nasser~\etal~proposed a UAV that can perform human following and gesture recognition with an onboard Xtion depth camera~\cite{uav_hri-human-tracking_naseer_13}. More recently, Yao~\etal~integrated a face detector and a feature tracker in order to achieve robust human tracking with a miniature robotic blimp~\cite{uav_hri-human-tracking_yao_17}. Note that these systems are not able to track and follow the user robustly in every situation, e.g., when the user is occluded by other objects/people. In order to tackle the occlusion problem, Hepp~\etal~presented a human-following system based on ultra-wideband (UWB) radio and released their implementation as open-source software~\cite{companion_uav-hepp-iros16}.

A few works on human-following UAV focus on filming. Huang~\etal~combined a stereo camera-based human following method with a dynamic planning strategy to film a human action in a more expressive manner~\cite{hdi_huang_icra18}. Zhou~\etal~designed a flying drone that could keep tracking a human motion using a normal color camera~\cite{hdi_zhou_icra18}. Bentz~\etal~presented an assistive aerial robot that could observe regions most interesting to the human and broadcast these views to the human’s augmented reality display~\cite{hdi_bentz_icra19}. This resulted in reduced head motions of the human as well as improved reaction time.

%In addition to research UAVs, commercial UAVs such as Skydio Camera~\cite{uav_product_skydio-camera}, DJI Phantom 4~\cite{uav_product_dji-phantom-4}, DJI Spark~\cite{uav_product_dji-spark}, and Hover Camera~\cite{uav_product_hover-camera} also demonstrate human following with computer vision and machine learning techniques.\footnote{Since there is no scientific publication, we derive our conjectures from product reviews and online videos.}

\subsection{User Interfaces for UAVs}
\label{subsec:user_interfaces_for_uavs}

%This subsection lists up the control interfaces that enable users to control UAVs more naturally with hand gestures, facial expression, or voice commands.

In 2013, Monajjemi~\etal~presented a method to command a team of UAVs by using face and hand gestures~\cite{uav_hri-interface_monajjemi_13}. Later, Monajjemi~\etal~extended their work by commanding a team of two UAVs using not only face engagement and hand gestures but also voice and touch interfaces~\cite{uav_hri-interface_monajjemi_14}.
Similar to Monajjemi's works on multi-modal interaction above, MohaimenianPour \& Vaughan~\cite{hdi_pour_iros18} and Nagi~\etal~\cite{hdi_nagi_hri14} realized UAV control with hands and faces by relying on visual object detectors and simple preset rules

Unlike Monajjemi's works on multi-modal interaction above, Sun~\etal~focused on piloting a drone with gesture recognition by combining a visual tracker with a skin pixel detector for robust performance~\cite{uav_hri-interface_sun_17}.
Similarly, Lichtenstern~\etal~demonstrated a system where a user can control multiple UAVs using hand gestures~\cite{uav_hri-interface_lichtenstern_12}.

Constante~\etal~aimed to improve the hand gesture interface of UAVs by proposing a new algorithm transfer learning algorithm that can exploit both online generic and user-specific hand gestures data~\cite{uav_hri-interface_costante_14}. Burke \& Lasenby presented a very fast and simple classification method to control a UAV with pantomimic gestures, in which the main idea to use a gesture that is similar to the desired action of UAV as a gesture command~\cite{uav_hri-interface_burke_15}.

More recently, Bruce~\etal~proposed the use of facial expression for 3D trajectory control of UAVs~\cite{facial_expression-bruce-crv17}.
%For example, the UAV can be signalled to initiate a boomerang trajectory motion by utilizing the learned facial expression. Interestingly, Bruce~\etal~used the analogy of shooting a bow to control the ``shooting power'' of the drone. In shooting a bow, the farther the string is pulled back, the larger the shooting power. In ``shooting'' a UAV, Bruce~\etal~utilized the distance of the face from the UAV (the farther the face is from the UAV, the smaller the face appears in the image) to control the shooting power of the UAV.
Previously, we have also demonstrated a drone that could react to the user's facial expression~\cite{facial_expression-liew-jface12}. In contrast to facial expression, Huang~\etal~directed a UAV in a known environment via natural language commands~\cite{uav_hri-interface_huang_10}.
\section{From Remote-Controlled to Social UAVs}
\label{sec:from_remote_controlled_to_social_uavs}
% ========== ========== ==========

This section summarizes efforts in sociability studies for realizing companion UAVs. This section also offers different perspective from the recent survey work of social drones~\cite{social-interaction_baytas_chi19}. We first discuss the social perception of UAVs, followed by topics on emotion and intention expression of UAVs through motions, lights, or displays. Then, we briefly describe related works in gesture interaction and physical interaction with UAVs.

\subsection{Social Perception of UAVs}
\label{subsec:social_perception_of_uavs}

Designing companion UAVs which invite social interaction is important. Wojciechowska~\etal~investigated the best way for a flying robot to approach a person~\cite{social-interaction_wojciechowska_hri19}. Yeh~\etal~found that a drone with circular body shape, face, and voice could reduce the proximate distance between a social drone and the user~\cite{social-interaction_yeh_hai17}. In addition, there are also studies on user perception on UAV, focusing on assistance during emergency situations~\cite{hdi_khan_chi19}, privacy and security issues~\cite{hdi_chang_chi17}, and autonomous behaviors~\cite{hdi_nowacka_ubicomp15}. 

Different from the social interaction works mentioned above, Abtahi~\etal~explored the touch interaction in HDI and participants preferred interacting with a safe-to-touch drone in the studies~\cite{hdi_abtahi_imwut17}. In particular, users feel safer and were less mentally demanding when interact with the safe-to-touch drone.

\subsection{Emotion and Intention Expression of UAVs}
\label{subsec:emotion_and_intention_expression_of_uavs}

Dancers use various kinds of motion to express their emotions. Sharma~\etal~used Laban motion analysis (a common method used by artists to express emotions) for UAVs to express their affective states~\cite{uav_hri-social-study_sharma_13}.
%In the experiments, all eighteen participants liked the idea of robots expressing their emotions through affective motion and perceived most of the expressed affects correctly. 
Aiming to deliver an opera performance, Eriksson~\etal~also described their method of designing expressive motions interactively with a choreographer for drones~\cite{hdi_eriksson_chi19}.

Similarly, Cauchard~\etal~presented a model for UAVs to express emotions via movements~\cite{uav_hri-social-study_cauchard_16},
%They proposed a set of eight personalities (emotional states) and designed the corresponding UAV's movements. For example, a \textit{brave} UAV would fly quickly, look at a person directly, and never go backwards. A \textit{sleepy} UAV would fly slower, stay low, and respond to commands with delays.
believing that encoding these emotions into movements could help users to comprehend the UAV's internal states.
%One good example is that, without looking into the controller screen, the users can identify a UAV's state of tiredness (\textit{low battery} or in \textit{sleepy} emotional state) intuitively when the UAV starts to move sluggishly and fly low.

In contrast to emotion expression, Szafir~\etal~used the UAV's motion to express the robot's intention~\cite{uav_hri-social-study_szafir_14}.
%They designed a few motion primitives for UAVs like flying along a more natural arc path and showed that eighty-five participants perceive the robot's intention better. 
Walker~\etal~expand this work by visualizing robot motion intent using an augmented reality technique~\cite{hdi_walker_hri18}. Colley~\etal~also investigated drone motion as direct guidance for pedestrians rather than equipping drones with a display or indicators~\cite{hdi_colley_mum17}.

Duncan~\etal~have similar idea and presented an initial study for UAVs to communicate their internal states to bystanders via flying patterns~\cite{hdi_duncan_icra18}. In their seminal work, Firestone~\etal~performed a participatory design with users for UAVs to communicate internal states effectively via flying patterns~\cite{hdi_firestone_hri19}.

%Malliaraki presented an online video regarding the concept of natural HDI~\cite{companion_uav-malliaraki-vimeo17}. In particular, she designed a drone system that can convey its intentions or give feedback to a person by programming the drone's position, orientation, flying speed, and flying direction (similar to the motion expression work~\cite{uav_hri-social-study_sharma_13}). Interestingly, she also mentioned about the control of the drone using gestures (similar to the gesture interface works~\cite{uav_hri-interface_monajjemi_15, uav_hri-interface_sun_17, uav_hri-interface_costante_14, uav_hri-interface_burke_15}) and facial expression (similar to the facial expression works~\cite{facial_expression-liew-jface12, facial_expression-bruce-crv17}). 

LED light has also been used for UAVs to express their emotion and intent. Arroyo~\etal~described a social UAV that performs four different expressions with head movement and two color LED eyes~\cite{uav_hri-social-study_arroyo_14}. Szafir~\etal~also used a ring of sixty-four color LEDs as a reliable cue for the UAV to convey intention to the user~\cite{light_interaction-szafir-hri15}.
%In particular, they found success in a \textit{gaze} signaling pattern of the ring LEDs, where two regions of LEDs (reminiscent of human eyes, with the same size and same interocular distance) are lit up toward a certain direction where the robot intends to fly to.

Instead of using LED light, some works rely on displays or projectors to convey information to users, including a small drone with an OLED display for telepresence function HDI~\cite{display_interaction-gomes-chi16}, a flying display system for crowd control during emergency situations~\cite{display_interaction-schneegass-chi14}, a flying projector-screen system with two UAVs~\cite{display_interaction-nozaki-chi14}, and flying UAVs with onboard projectors for social group interactions~\cite{hdi_scheible_perdis13}, interactive map application~\cite{hdi_brock_perdis18}, navigation guidance~\cite{hdi_knierim_chi18}, and gesture interaction \cite{hdi_cauchard_hri19}.

\subsection{Gesture Interaction with UAVs}
\label{subsec:gesture_interaction_with_uavs}

Inspired by human interaction with birds, Ng \& Sharlin studied the effectiveness of a few hand gestures in commanding a UAV~\cite{uav_hri-social-study_ng_11}. Participants were very engaged when having gesture interaction with the UAV and spoke to the UAV like a pet.
%Besides, they also found that the participants liked the \textit{stop} and \textit{come} gestures the most, as these gestures are common and often used.
Cauchard~\etal~also performed similar Wizard-of-Oz experiments and most participants interacted with the UAV as if it were a pet~\cite{uav_hri-social-study_cauchard_15}.
%Several participants mentioned feeling attached to the UAVs. The authors also observed that out of the 216 unique interactions that were found, 96 interactions are related to body or hand gestures and 59 interactions are related to sound commands.
%Among these interactions, the most commonly used gestures and sound commands are \textit{fly closer}, \textit{stop by me}, \textit{follow me}, \textit{fly sideways}, \textit{fly higher/lower}, \textit{fly to a precise location}, \textit{get attention}, and \textit{take a picture of an object}.
%After the experiments, several participants mentioned that they would like an interface for emergency landing in case anything goes wrong. In the conclusion, the authors also share their belief that the idea of a companion UAV will be realized as UAVs become smaller and quieter. 
E~\etal~later expand this experiment in different culture setting and found similar results~\cite{uav_hri-social-study_e_17}.

Aiming to increase the naturalness of HDI, Peshkova \etal~surveyed gesture interaction techniques that have been applied for UAV control based on three mental models: the imitative class (controls the UAV motions with the user's body movements), the instrumented class (controls the UAV motions with a physical controller or an imaginary object), and the intelligent class (interacts with a UAV as if the UAV is an intelligent agent)~\cite{gesture_interaction-peshkova-pc17}.

On the other hand, Pfeil~\etal~studied the effectiveness of different interaction techniques of the upper body in UAV control~\cite{gesture_interaction-pfeil-iui13} (including all the three interaction classes mentioned by Peshkova~\etal~\cite{gesture_interaction-peshkova-pc17}). They found that the \textit{proxy} technique, in which the user moves the UAV as he/she is grasping the UAV in his/her hand, is the best out of the five developed interaction techniques.

\subsection{Physical Interaction with UAVs}
\label{subsec:physical_interaction_with_uavs}

Physical interaction is rare in the UAV literature compared to gesture interaction. Knierim~\etal~used physical interaction with a flying robot as a novel input (touch/drag the flying robot) and output (the flying robot generates forces feedback) modalities for a user interface~\cite{hdi_knierim_tei18}.
Abtahi~\etal~proposed a haptic interaction\footnote{Physical HDI with a virtual reality display in their context.} system, where an actual UAV is used to enhance user's physical perception in virtual reality environment~\cite{hdi_abtahi_chi19}.
Soto~\etal~explored the idea of using a leashed UAV as a navigator to guide visually impaired people~\cite{hdi_soto_assets17}. 
%Imagine we can walk with our UAVs (like pets) with a leash/wire in the near future, tethered robotics is also closely related to physical HDI.\footnote{The tether could be wire, serves the purpose of a normal leash and offer a few additional merits: used for power transmission (longer flight time) and for video transmission (more secure and faster).} 
%Mention [DroneNavigator: Using Leashed and Free-Floating Quadcopters to Navigate Visually Impaired Travelers (ASSETS '17)]~\cite{hdi_soto_assets17}

%~\cite{tethered_robotics-fotokite-16}.
%Readers who are interested in more details on the technical aspect of tethered robotics can refer to the related papers in our survey list (more details in Appendix~\ref{appendix:uav_database_update_and_online_sharing}). For example, Galea \& Kry developed an interesting tethered UAV for stippling (i.e., drawing with different sizes of dots)~\cite{tethered_robotics-galea-iros17}. Lupashin \& D'Andrea also designed a tethered UAV that can hover stably by using inertial sensing~\cite{tethered_robotics-lupashin-iros13}.

% ========== ========== ==========
\section{Observation and Lessons Learned}
\label{sec:observation_and_lessons_learned}
% ========== ========== ==========

Throughout the review process and personal experience, we noticed several patterns in the literature and learned a few lessons in designing companion UAVs. We discuss these observation (including ideas exploration) in this section, including: (i) UAV form, (ii) appearance design, (iii) integrated human-accompanying model, (iv) integrated human-sensing interface, (v) safety concerns, (vi) noise issue and sound design, and tactile interaction. Note that several aspects mentioned in this section could be potentially improved by drawing inspiration from the human-computer interaction or human-robot interaction literature. In next section, we will present a more concise guidelines and recommendations towards realizing companion UAVs.

\subsection{UAV Form}
\label{subsec:uav_form}

Almost all papers considered in this work use multirotor UAV as a platform to demonstrate UAV flight or to carry out social experiments. From our long experience working with drones, we agree that multirotors are more convenient for experiments (in term of availability) and presentation (in term of flight quality) but their noise level is too annoying for companion UAVs.
%Previously, we also carried out a preliminary online study with forty-two participants and found that users preferred a penguin-shaped blimp (nature-inspired design) over a ball-shaped multirotor (functional design)~\cite{uav_hri-design_liew_13}.
% It was expected that more participants would prefer the ball-shaped rotorcraft as it is more useful and possesses unique characteristics needed to excel in a disaster situation (which was also explained to the participants). The penguin-shaped blimp has slower motion, but has a charming appearance and is able to ``swim'' quietly and naturally in the air. This finding is coherent with the design concept proposed by human-computer interaction (HCI) researchers~\cite{uav_hri-design_jung_10}, suggesting that appearance is an important design factor for companion UAVs.

We argue that a blimp or balloon type UAV is more suitable as a form for companion UAVs. We list up two technical blimp design papers that could be an alternative form of companion UAVs. First, Song~\etal~used a novel idea to hide the propulsion unit in the center of a blimp and designed a blimp that is safe to touch and interact with~\cite{hdi_song_iros18}. Second, Yamada~\etal~designed a blimp with micro-blower with no rotating blade (hence safer and quieter)~\cite{hdi_yamada_chi19}. It is also worth mentioned that Drew~\etal~designed a small flying robot using electro-hydrodynamic thrust with no moving part but it is tiny and cannot handle large payload.~\cite{hdi_drew_ral18}

\subsection{Appearance Design}
\label{subsec:appearance_design}

Appearance design of drones is important as the design affect users' perception~\cite{social-interaction_wojciechowska_imwut19}. A few HDI studies mentioned in Section~\ref{subsec:social_perception_of_uavs} investigated the user perception on drones. For example, it is found that a round shape flying robot has higher social acceptance~\cite{social-interaction_yeh_hai17} and emergency response drones with a prominent appearance that is easily distinguishable from recreational drones can gain user trust~\cite{hdi_khan_chi19}. HDI researchers have mentioned about the importance of color in drone design~\cite{hdi_chang_chi17}. At the time of writing, no study has investigated the color design in companion UAVs and the most related study we can find is using color of balloons to visualize their surrounding air quality~\cite{hdi_kuznetsov_ubicomp11}.

\subsection{Human Accompanying Model}
\label{subsec:human_accompanying_model}

A few papers demonstrated human-following capability of companion UAVs (Section~\ref{subsec:human_following_uavs}) and Wojciechowska \etal~investigated the best way a flying robot should approach a person~\cite{social-interaction_wojciechowska_hri19}. We noticed that there is a lack of a general model to unify various human accompanying behaviors of companion UAVs, including approaching, following, side-by-side walking, leading or guiding, and flying above the user (to help observing things far away). This observation is also applicable to ground robots. With a unified human accompanying model, companion UAVs are expected to be able to truly accompany and interact with a human more naturally. For more details, we have summarized related works of various human accompanying modes of both flying robots and mobile robots in our previous work~\cite{liew_dissertation_16}.

\subsection{Human Sensing Interface}
\label{subsec:human_sensing_interface}

Human can achieve natural interaction with each other using face, gesture, touch, and voice modalities simultaneously. A few papers demonstrated HDI with multiple modalities but most papers focus on a single modality. It is not straightforward to realize a companion UAV with all modalities since researches/engineers often focus on methods with a single modality. Recently, great effort in the deep learning topic has led to a more integrated human sensing interface, such as the OpenPose library integrates visual tracking of human body, face, and hand simultaneously~\cite{hdi_cao_arxiv19}. We should leverage these powerful tools more in order to extent companion UAV research. As a good example, Huang~\etal~utilized the OpenPose library for human-following and realize a high-level autonomous filming function with their UAV~\cite{hdi_huang_icra18}. A standard human sensing interface (could be updated regularly) is crucial in facilitating a HDI study, not only it can accelerate HDI progress, but also make comparison study more effective.

\subsection{Safety Concerns}
\label{subsec:safety_concerns}

HDI safety is an important design aspects of companion UAV.
%According to a pilot study, UAVs with a mass over 1.2 kg flying towards a human could result in head and neck injuries~\cite{uav_hri-safety_campolettano_17}.
Most UAVs described in this work have sharp rotating propellers that could injure nearby human, especially human eyes---such accidents have been observed in a medical case report~\cite{uav_hri-safety_moskowitz_18} and a formal news~\cite{uav_hri_safety-drone_accident}. Existing solutions\footnote{Including commercial UAV examples.} include using ring-shape protectors around the propellers~\cite{uav_product_ar-drone, uav_product_aibot-x6}, having net cases that fully cover the propellers~\cite{hdi_abtahi_imwut17, hdi_salaan_ral19, uav_product_hover-camera, uav_product_snap}, or designing a cage to cover the entire drone~\cite{hdi_abtahi_chi19, challenges-briod-jfr13, uav-background_coaxial_briod_iros13, challenges-kornatowski-iros17, uav_product_flyability, uav_product_fleye} but these modifications worsen the flight efficiency and shorten the flight time (due to the increased payloads).
%As an alternative, using a smaller multirotor could be less harmful to the nearby humans, but they would have very limited payloads (hence, lower sensing capability and smaller battery with shorter flight time).
%Using a blimp UAV could achieve a safer HDI but compared to a multirotor, a blimp has some drawbacks such as slow flight response and susceptibility to wind.
%Designing a small flapping-wing UAV like a bird or a bee is another potential solution. However, the flight dynamics of a bird or a bee are complex, and a flapping-wing UAV normally cannot handle large payloads and cannot have much onboard processing capabilities. 

%[Beyond The Force: Using Quadcopters to Appropriate Objects and the Environment for Haptics in Virtual Reality (CHI'19)] has a mini cage design~\cite{hdi_abtahi_chi19}
%[Development and Experimental Validation of Aerial Vehicle With Passive Rotating Shell on Each Rotor (RAL'19)] has covers for each propeller~\cite{hdi_salaan_ral19}
%[Drone Near Me - Exploring Touch-Based Human-Drone Interaction (IMWUT'17)] has covers for each propeller~\cite{hdi_abtahi_imwut17}

Recently, Lee~\etal~claimed that a Coanda UAV has several advantages over standard UAVs, such as crash resistance and flight safety, thanks to its unique mechanical design~\cite{uav_hri-design_lee_17}. A UAV with flexible structures could be less harmful when it unavoidably hit a user (as the UAV structure will absorb the crash impact). Based on this idea, UAVs with soft body frame~\cite{safety_concerns-mintchev-ral17} and flexible propellers~\cite{hdi_jang_ral19} have been proposed.

%[Design and Experimental Study of Dragonfly-Inspired Flexible Blade to Improve Safety of Drones (RAL'19)] flexible blade design, about 1/10 of impact force compared to the original blade~\cite{hdi_jang_ral19}
%[The Safety Rotor—An Electromechanical Rotor Safety System for Drones (RAL'18)], potential improvement, proximity touch sensing.

In addition to the physical safety, making users feel safer (less cognitive burden) is also important. From our experience, most users are afraid that the UAV is going to crash when it starts to move (because unlike a common ground vehicle, a conventional UAV has to tilt or roll in order to move). We then designed a special type of drone---a holonomic UAV, where it can move horizontally without tilting, and the users expressed that the holonomic flight makes them feel safer. While several holonomic UAVs exists,\footnote{Papers can be found by searching the ``holonomic'' keyword in the UAV paper list mentioned in Appendix~\ref{appendix:uav_database_update_and_online_sharing}} there is no formal HDI study of holonomic UAV so far to the best of our knowledge.

\subsection{Noise Issue and Sound Design}
\label{subsec:noise_issue_and_sound_design}

In addition to the safety concerns, noise issue is also important for HDI. Most UAVs produce unwanted noise with their high speed and high power rotating propellers. In our test, the noise of a commercial UAV~\cite{uav_product_ar-drone-2} was measured to be as high as 82 dB one meter away, and is very close to the hazardous level of 85 dB as legislated by most countries~\cite{uav_hri-design_noise-issue-1}. Studies also suggested that noise has a strong association with health issues~\cite{uav_hri-design_noise-issue-2} and increased risk of accidents~\cite{uav_hri-design_noise-issue-3}. These findings suggest that noise issue should be seriously considered in HDI.
%Interestingly, Islam~\etal~pointed out that UAV noise causes discomfort to elephants when UAVs are used in wildlife monitoring applications~\cite{noise_consideration-islam-duke17}, implying that the noise issue deserves serious attention for a more effective animal-drone interaction as well. Compared to rotor-type UAVs, blimp or balloon-type UAVs could be a better platform for companion UAVs because of their lower noise level.

Sound design~\cite{sound_interaction-lyon-jas00} is also a related topic for HDI. For example, car manufacturers harmonically tune the engines' noise so that their cars can sound more comfortable to the users~\cite{sound_interaction-kim-jpc17}. Norman discussed about the emotional association of sounds and everyday products in his book~\cite{sound_interaction-norman-book13}. For example, an expensive melodious kettle (when the water is boiling) and a Segway self-balancing scooter (when the motor is rotating) sound exactly two musical octaves apart, making the unwanted noises sound like music. In the robotics context, Moore~\etal~utilized the motor noise to facilitate user interaction~\cite{sound_interaction-pena-hri17} and Song \& Yamada proposed to express emotions through sound, color, and vibrations~\cite{uav_hri-design_song_17}.
%Moreover, we have noted that DJI incorporates this kind of sound design in its latest commercial drones~\cite{uav_product_dji-phantom}. Specifically, during takeoff, a DJI drone would purposely rotate its propeller with a few cycles of high and low power within a short period of time, to signify that the drone is going to takeoff. To the best of our knowledge, there is no formal study regarding the sound design of UAV.

The use of non-vocal, non-verbal, or non-linguistic utterance (NLU) for HDI is also a potential way to enhance a UAV's characteristics and expressiveness. It should be noted that NLU might be more useful than speech during a HRI flight as the propeller noise make speech recognition difficult. In movies, robots such as R2-D2 also use non-verbal utterance to enhance its communication with characters.
%In the HCI context, Brown~\etal~described the usefulness of NLU and vibration in a remote interpersonal communication~\cite{thermal_interaction-brown-chi09}. In the HRI context, Read \& Belpaeme proposed to use NLU for child-robot interaction~\cite{sound_interaction-read-hri12}. In a later study, Robin \& Belpaeme also found that users tend to prefer a robot that combines NLU and natural language, compared to a robot that only uses either NLU or natural language~\cite{sound_interaction-read-hri14}. It is worth mentioning that in addition to enhancing a UAV's characteristics and expressiveness, NLU is also useful for humans to localize a robot~\cite{sound_interaction-cha-hri18}. In particular, Cha~\etal~found that broadband sounds (containing a larger range of frequencies) were preferred by 24 participants in their studies as the broadband sounds are less annoying than tonal sounds (containing a small range of frequencies, such as a traditional alarm)~\cite{sound_interaction-cha-hri18}.
For more details on the study of NLU on HRI, readers are recommended to read the PhD thesis of Read~\cite{sound_interaction-read-dissertation14}.

\subsection[Tactile Interaction]{Tactile Interaction\footnote{We focus on thermal and touch interactions here, which have subtle difference with physical interaction (involves force feedback) mentioned in Section~\ref{subsec:physical_interaction_with_uavs}.}}
\label{subsec:tactile_interaction}

Pe{\~n}a \& Tanaka proposed the use of a robot's body temperature to express emotional state~\cite{thermal_interaction-pena-hri18}.
%Thermal sensation has the characteristic of privacy, where the robot can express its emotion to the interacting person without others noticing it.
Park \& Lee also studied the effect of temperature with a companion dinosaur robot and found that skin temperature significantly affects users' perception~\cite{thermal_interaction-park-ro14}.
%In the HCI context, Lee \& Lim identified the key expression elements of a heat source, including the temperature, duration, location, and temperature change rate~\cite{thermal_interaction-lee-puc12}.
It would be interesting and useful to explore this new thermal application area. For example, companion UAVs with warm/hot body frame could signify a status of hardworking/exhaustion.

Equipping UAVs with touch sensing capability allows richer HDI. For instance, a UAV could perceive a person's love if it could sense the person's gentle stroke. Prior studies and results in the social robotics field~\cite{hri_yohanan_ijsr12} could be integrated into a UAV to realize a more personalized companion UAV. In addition, from the engineering perspective, a UAV with touch sensors on its propeller guard could also sense a nearby human instantly and enhance HDI safety.

%\subsection{Swarm Interaction}
%\label{subsec:swarm_interaction}

%Most companion UAV works that we have reviewed so far focus on a single UAV. When the autonomous and social capabilities of UAV become more mature, the idea of multiple companion UAVs could opens up many new opportunities. For example, multiple companion UAVs could share their camera views and have a better understanding of the user and the environment. Besides, multiple companion UAVs could act like a GPS system and co-locate a user in a 3D space more accurately (and therefore perform a human following function more accurately). This topic is challenging and some open questions are, e.g., how can a user control multiple UAVs simultaneous without high cognitive loads, how should multiple UAVs communicate their intentions to the user, should the UAV swarm consists of one type of homogeneous UAVs or several types of different UAVs like multirotors and blimps, etc. Monajjemi~\etal~tackled the first open question by designing several multi-modal NUI for proximate interaction with multiple UAVs~\cite{uav_hri-interface_monajjemi_13, uav_hri-interface_monajjemi_14} but more works (and more opportunities) remain to be explored. Focusing on human-system interface and human factors concerns, Hocraffer \& Nam performed a meta-analysis of 27 UAV swarm management interfaces~\cite{uav_swarm-hocraffer-ae17}. Nevertheless, they focused on teleoperated UAVs (remote-controlled interaction) and did not consider proximate interaction in their review.

% ========== ========== ==========
\section{Guidelines and Recommendations}
\label{sec:guidelines_and_recommendations}
% ========== ========== ==========

%Should summarize key guidelines and recommendations concisely from the previous section: (i) UAV form, (ii) appearance design, (iii) integrated human-accompanying model, (iv) integrated human-sensing interface, (v) safety concerns, (vi) noise issue and sound design, and tactile modalities.

After mentioning several observation in the companion UAV literature (and our experience in designing companion UAVs), this section presents several design and research recommendations for companion UAVs. Note that almost all the topics discussed below are applicable to both the engineering development and social interaction experiments.

First, in the topic of UAV form and appearance design, we recommend two kind of platforms in the engineering or social experiments: (i) a palm-sized multirotor UAV with cage design (e.g.,~\cite{hdi_abtahi_chi19}) for agile and accurate motion, safer interaction (less impact and less chance to get hurt by the propellers), affordance that invites touch interaction (e.g.,~\cite{hdi_abtahi_imwut17}), if noise and flight time are not a big issue; (ii) a hugging-sized blimp with hiding propellers (e.g.,~\cite{hdi_song_iros18}) or novel propulsion unit with no rotating part (e.g.,~\cite{hdi_yamada_chi19}) for quieter and calm interaction, safer interaction, and longer interaction time, if agile and accurate response are not a big issue.

Second, we suggest to pay more attention to integrated human-accompanying models and human-sensing interfaces in order to support a more realistic HDI. Human-accompanying model should integrate functions of human approaching, following, leading, side-by-side walking, bird-eye viewing on the top for a more natural HDI. Similarly, human-sensing interface should integrate at least four modalities of human tracking, hand tracking, face tracking, and voice interaction (e.g.,~\cite{uav_hri-interface_monajjemi_14}). In the engineering field, UAVs should also perform environment sensing at the same time so that they can accompany their users without hitting obstacle (e.g.,~\cite{uav_hri-human-tracking_lim_15}).

Third, more related to the social interaction studies, we recommend to explore the ideas of sound design, tactile interaction, and holonomic flight motions of UAV. When individual interaction becomes more mature, we should try integrating the visual and audio expression, gesture and physical and tactile interactions, and investigate the long term HDI.

Fourth, we also encourage HDI researchers to share development code among companion UAV studies to facilitate comparison study under a shared metrics. In addition to our recommendation, one could also draw inspiration from the practices and know-how in the aerospace field~\cite{air_traffic_control-hutchins-ssrr07} and social robotics field~\cite{hri_coeckelbergh_ijsr09, hri_harper_ijsr09}.

\begin{figure}[tb]
  \centering
  \includegraphics[trim={3.8cm 8cm 17.5cm 7cm},clip,height=3.4cm]{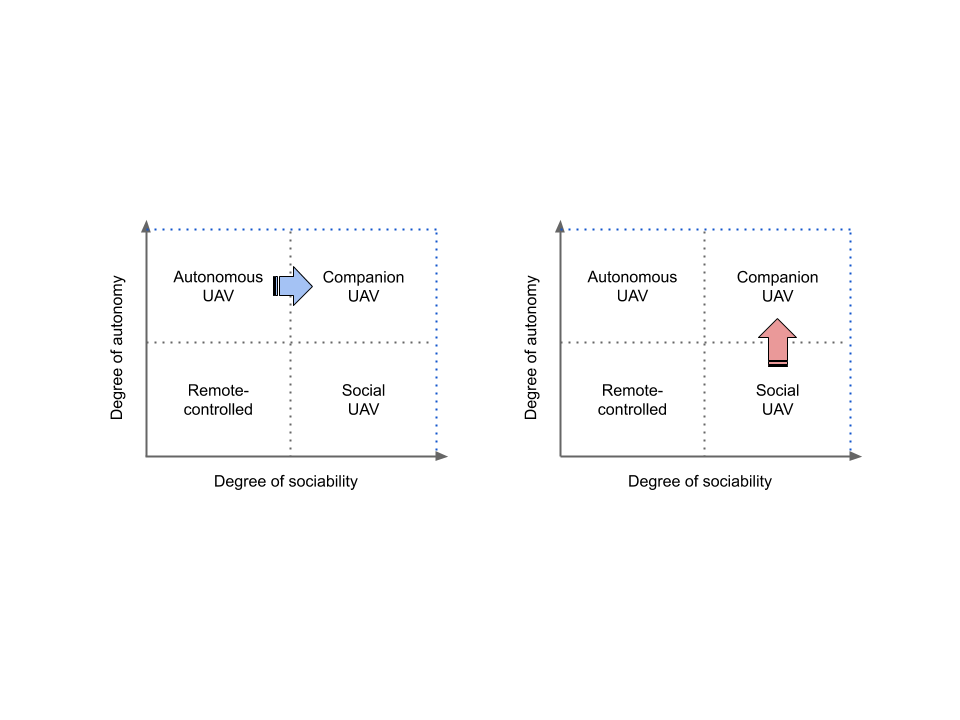}
  \includegraphics[trim={18.5cm 8cm 3.5cm 7cm},clip,height=3.4cm]{uavPerceptualMapGuideline.png}
  \caption{Towards companion UAV from the autonomous UAV (left) and social UAV (right) categories.}
  \label{fig:uav_perceptual_map_guideline}
\end{figure}

Fifth, last but not least, we suggest engineering research to (i) incorporate findings in the HDI studies (such as accompanying a user with a proximate distance that is comfortable to the user) into their technical development, and (ii) perform HDI study after developing a new function in order to confirm its usefulness and social acceptance (efforts from autonomous UAV to companion UAV, corresponding to the Fig.~\ref{fig:uav_perceptual_map_guideline} (left)). At the same time, we suggest HDI studies to perform experiments with a real drone integrated with autonomous capabilities (such as accompanying a person and avoid obstacles autonomously) in order to deal with HDI studies in a more realistic scenario (efforts from social UAV to companion UAV, corresponding to the Fig.~\ref{fig:uav_perceptual_map_guideline} (right)).

%5. Long term interaction, tethered flight, bird falconeering.
%6. Policy making, such as electronic vehicle plate where the UAV's info can be identified by a smartphone or smartwatch, making users more comfortable.

% ========== ========== ==========
\section{Conclusion}
\label{sec:conclusion}
% ========== ========== ==========

Technological advancements in small-scale UAVs have led to a new class of companion robots---a companion UAV. After identifying and coordinating related works from the UAV's autonomy and sociability perspectives, we found that recent research efforts in companion UAVs focus on a single issue, either an engineering or a social interaction issue, rather than having an ambitious goal to tackle both issues in a paper. While this might be the nature of research (i.e., specialize on a topic), we encourage future works to emphasis on both aspects of companion UAVs as the integration of these two interrelated aspects is essential for an effective HDI.

We also list up our observation throughout this review and propose guidelines to perform companion UAV designs and researches in the future. In addition to individual topics such as engineering functions and social interaction studies with new modality, we argue the importance of devising an integrated human-accompanying model and an integrated human-sensing interface to advance the development of companion UAVs. We also suggest researchers to share the programming codes used in their experiments to facilitate comparison study and consolidate findings in companion UAV works.

Most of the related papers focus on the development of the UAV itself. In contrast, it is also feasible to design an environment that could enable an easier navigation of companion UAVs. Public issues~\cite{public_issue-pytlikZillig-tsm18} and policy making~\cite{public_issue-un-ocha14} are also important for facilitating a social acceptance of UAVs. Lastly, while an affective model is an important aspect of companion UAVs, we argue that it is beyond the scope of this paper. We believe that affective models developed for general companion robots are applicable to companion UAVs when companion UAVs have more mature and integrated autonomous functions and socially interactive capabilities.

% ========== ========== ==========
\section*{Compliance with Ethical Standards}
\label{sec:compliance_with_ethical_standards}
% ========== ========== ==========

\textbf{Conflict of interest} The authors declare that they have no conflict of
interest.

% ========== ========== ==========
\appendix
% ========== ========== ==========

% ========== ========== ==========
\section{UAV Database Update and Online Sharing}
\label{appendix:uav_database_update_and_online_sharing}
% ========== ========== ==========

The UAV database is shared online via Google Sheets (\url{https://tinyurl.com/drone-paper-list}). In the tables (one table per year), we list all the UAV-related papers from top journals/conferences since 2001 along with their related topics, abstracts, and details such as hardware summary. %Note that this list is being updated twice a year and is growing. 
This list is particularly useful to: (i) search related works on a particular topic in UAV, e.g., HRI; (ii) search related works on a particular type in UAV, e.g., blimp; (iii) search related works on a particular UAV platform, etc. We believe that this list is not only beneficial for newcomers to the UAV field but also convenient for experienced researchers to cite and compare related works.

In addition to Google Sheets, we also use an open-source file tagging and organization software~\cite{online_application_tagspaces}. TagSpaces enables readers to search papers with multiple tags or/and keywords effectively. Moreover, since original papers (PDF files) cannot be shared with readers due to copyright issues, for each paper entry, we create an HTML file that contains public information (such as abstract, keywords, country, paper URL link, and video URL link) for easier reference. To setup TagSpaces and download all the HTML files, please refer to our website at \url{https://sites.google.com/view/drone-survey}.

\bibliographystyle{spmpsci}      % mathematics and physical sciences
\bibliography{bib_jun,bib_jun2}

\begin{thebibliography}{100}
\providecommand{\url}[1]{{#1}}
\providecommand{\urlprefix}{URL }
\expandafter\ifx\csname urlstyle\endcsname\relax
  \providecommand{\doi}[1]{DOI~\discretionary{}{}{}#1}\else
  \providecommand{\doi}{DOI~\discretionary{}{}{}\begingroup
  \urlstyle{rm}\Url}\fi

\bibitem{hdi_abtahi_chi19}
Abtahi, P., Landry, B., Yang, J.J., Pavone, M., Follmer, S., Landay, J.A.:
  Beyond the force: Using quadcopters to appropriate objects and the
  environment for haptics in virtual reality.
\newblock In: Proc. ACM Conference on Human Factors in Computing Systems (CHI),
  pp. 1--13 (2019)

\bibitem{hdi_abtahi_imwut17}
Abtahi, P., Zhao, D.Y., E., J.L., Landay, J.A.: Drone near me: Exploring
  touch-based human-drone interaction.
\newblock Proc. ACM Interactive, Mobile, Wearable and Ubiquitous Technologies
  (IMWUT) \textbf{1}(3), 813--824 (2017)

\bibitem{social-interaction_agrawal_ubicomp15}
Agrawal, H., won Leigh, S., Maes, P.: L'evolved: Autonomous and ubiquitous
  utilities as smart agents.
\newblock In: Proc. ACM International Joint Conference on Pervasive and
  Ubiquitous Computing (UbiComp), pp. 487--491 (2015)

\bibitem{uav_product_aibot-x6}
{Aibot X6} (2016).
\newblock \urlprefix\url{https://www.aibotix.com}

\bibitem{uav_hri-social-study_arroyo_14}
Arroyo, D., Lucho, C., Roncal, J., Cuellar, F.: Daedalus: {A} s{UAV} for social
  environments.
\newblock In: Proc. ACM/IEEE International Conference on Human-Robot
  Interaction (HRI), p.~99 (2014)

\bibitem{social-interaction_baytas_chi19}
Bayta{\c s}, M.A., {\c C}ay, D., Zhang, Y., Obaid, M., Yanta{\c c}, A.E.,
  Fjeld, M.: The design of social drones: A review of studies on autonomous
  flyers in inhabited environments.
\newblock In: Proc. ACM Conference on Human Factors in Computing Systems (CHI),
  pp. 1--13 (2019)

\bibitem{hdi_bentz_icra19}
Bentz, W., Dhanjal, S., Panagou, D.: Unsupervised learning of assistive camera
  views by an aerial co-robot in augmented reality multitasking environments.
\newblock In: Proc. IEEE International Conference on Robotics and Automation
  (ICRA), pp. 3003--3009 (2019)

\bibitem{uav_hri-design_noise-issue-2}
Bodin, T., Albin, M., Ard{\"o}, J., Stroh, E., {\"O}stergren, P., Bj{\"o}rk,
  J.: Road traffic noise and hyper tension: {R}esults from a cross-sectional
  public health survey in southern {S}weden.
\newblock Environmental Health \textbf{8} (2009)

\bibitem{uav-background_coaxial_briod_iros13}
Briod, A., Kornatowski, P., Klaptocz, A., Garnier, A., Pagnamenta, M.,
  Zufferey, J.C., Floreano, D.: Contact-based navigation for an autonomous
  flying robot.
\newblock In: Proc. IEEE/RSJ International Conference on Intelligent Robots and
  Systems (IROS), pp. 3987--3992 (2013)

\bibitem{challenges-briod-jfr13}
Briod, A., Kornatowski, P., Zufferey, J., Floreano, D.: A collision‐resilient
  flying robot.
\newblock Journal of Field Robotics \textbf{31}(4), 496--509 (2013)

\bibitem{hdi_brock_perdis18}
Brock, A.M., Chatain, J., Park, M., Fang, T., Hachet, M., Landay, J.A.,
  Cauchard, J.R.: Flymap: Interacting with maps projected from a drone.
\newblock In: Proc. ACM International Symposium on Pervasive Displays (PerDis),
  pp. 1--9 (2018)

\bibitem{facial_expression-bruce-crv17}
Bruce, J., Perron, J., Vaughani, R.: Ready—aim—fly! {H}ands-free face-based
  {HRI} for {3D} trajectory control of {UAV}s.
\newblock In: Proc. Conference on Computer and Robot Vision (CRV), pp. 307--313
  (2017)

\bibitem{uav_hri-interface_burke_15}
Burke, M., Lasenby, J.: Pantomimic gestures for human-robot interaction.
\newblock IEEE Transactions on Robotics (T-RO) \textbf{31}(5), 1225--1237
  (2015)

\bibitem{uav-paper_butzke_iros_15}
Butzke, J., Dornbush, A., Likhachev, M.: {3-D} exploration with an air-ground
  robotic system.
\newblock In: Proc. IEEE/RSJ International Conference on Intelligent Robots and
  Systems (IROS), pp. 3241--3248 (2015)

\bibitem{hdi_cao_arxiv19}
Cao, Z., Hidalgo, G., Simon, T., Wei, S.E., Sheikh, Y.: {OpenPose}: Realtime
  multi-person {2D} pose estimation using part affinity fields.
\newblock In: arXiv preprint arXiv:1812.08008, pp. 1--8 (2019)

\bibitem{uav_hri-social-study_cauchard_15}
Cauchard, J.R., E, J.L., Zhai, K.Y., Landay, J.A.: Drone \& me: An exploration
  into natural human-drone interaction.
\newblock In: Proc. ACM International Joint Conference on Pervasive and
  Ubiquitous Computing (UbiComp), pp. 361--365 (2015)

\bibitem{hdi_cauchard_hri19}
Cauchard, J.R., Tamkin, A., Wang, C.Y., Vink, L., Park, M., Fang, T., Landay,
  J.A.: Drone.io: A gestural and visual interface for human-drone interaction.
\newblock In: Proc. ACM/IEEE International Conference on Human-Robot
  Interaction (HRI), pp. 153--162 (2019)

\bibitem{uav_hri-social-study_cauchard_16}
Cauchard, J.R., Zhai, K.Y., Spadafora, M., Landay, J.A.: Emotion encoding in
  human-drone interaction.
\newblock In: Proc. ACM/IEEE International Conference on Human-Robot
  Interaction (HRI), pp. 263--270 (2016)

\bibitem{hdi_chang_chi17}
Chang, V., Chundury, P., Chetty, M.: Spiders in the sky: User perceptions of
  drones, privacy, and security.
\newblock In: Proc. ACM Conference on Human Factors in Computing Systems (CHI),
  pp. 6765--6776 (2017)

\bibitem{hri_coeckelbergh_ijsr09}
Coeckelbergh, M.: Personal robots, appearance, and human good: {A}
  methodological reflection on roboethics.
\newblock International Journal of Social Robotics (IJSR) \textbf{1}(3),
  217--221 (2009).
\newblock \doi{10.1007/s12369-009-0026-2}.
\newblock \urlprefix\url{https://doi.org/10.1007/s12369-009-0026-2}

\bibitem{hdi_colley_mum17}
Colley, A., Virtanen, L., Knierim, P., H{\"a}kkil{\"a}, J.: Investigating drone
  motion as pedestrian guidance.
\newblock In: Proc. International Conference on Mobile and Ubiquitous
  Multimedia (MUM), pp. 143--150 (2017)

\bibitem{uav_hri-platform_cooney_12}
Cooney, M., Zanlungo, F., Nishio, S., Ishiguro, H.: Designing a flying humanoid
  robot ({FHR}): {E}ffects of flight on interactive communication.
\newblock In: Proc. IEEE International Symposium on Robot and Human Interactive
  Communication (RO-MAN), pp. 364--371 (2012)

\bibitem{uav_hri-interface_costante_14}
Costante, G., Bellocchio, E., Valigi, P., Ricci, E.: Personalizing vision-based
  gestural interfaces for {HRI} with {UAV}s: {A} transfer learning approach.
\newblock In: Proc. IEEE/RSJ International Conference on Intelligent Robots and
  Systems (IROS), pp. 3319--3326 (2014)

\bibitem{uav-paper_das_ras_16}
Das, B., Couceiro, M.S., Vargas, P.A.: {MRoCS}: {A} new multi-robot
  communication system based on passive action recognition.
\newblock Robotics and Autonomous Systems (RAS) \textbf{82}, 46--60 (2016)

\bibitem{hdi_drew_ral18}
Drew, D.S., Lambert, N.O., Schindler, C.B., Pister, K.S.J.: Toward controlled
  flight of the ionocraft: A flying microrobot using electrohydrodynamic thrust
  with onboard sensing and no moving parts.
\newblock IEEE Robotics and Automation Letters (RA-L) \textbf{3}(4), 2807--2813
  (2018)

\bibitem{hdi_duncan_icra18}
Duncan, B.A., Beachly, E., Bevins, A., Elbaum, S., Detweiler, C.: Investigation
  of communicative flight paths for small unmanned aerial systems.
\newblock In: Proc. IEEE International Conference on Robotics and Automation
  (ICRA), pp. 602--609 (2018)

\bibitem{companion_uav-duncan-hri10}
Duncan, B.A., Murphy, R.R., Shell, D., Hopper, A.G.: A midsummer night’s
  dream: {S}ocial proof in {HRI}.
\newblock In: Proc. ACM/IEEE International Conference on Human-Robot
  Interaction (HRI), pp. 91--92 (2010)

\bibitem{uav_hri-social-study_e_17}
E, J.L., L.E, I., , Landay, J.A., Cauchard, J.R.: Drone \& wo: Cultural
  influences on human-drone interaction techniques.
\newblock In: Proc. ACM Conference on Human Factors in Computing Systems (CHI),
  pp. 6794--6799 (2017)

\bibitem{hdi_eriksson_chi19}
Eriksson, S., Unander-Scharin, {\AA}., Trichon, V., Unander-Scharin, C.,
  Kjellstr{\"o}m, H., H{\"o}{\"o}k, K.: Dancing with drones: Crafting novel
  artistic expressions through intercorporeality.
\newblock In: Proc. ACM Conference on Human Factors in Computing Systems (CHI),
  pp. 1--12 (2019)

\bibitem{faa_uav}
{Unmanned Aircraft Systems} (2017).
\newblock \urlprefix\url{https://www.faa.gov/uas/}

\bibitem{hdi_firestone_hri19}
Firestone, J.W., Qui{\~n}ones, R., Duncan, B.A.: Learning from users: An
  elicitation study and taxonomy for communicating small unmanned aerial system
  states through gestures.
\newblock In: Proc. ACM/IEEE International Conference on Human-Robot
  Interaction (HRI), pp. 163--171 (2019)

\bibitem{uav_product_fleye}
{Fleye flying robot} (2016).
\newblock \urlprefix\url{http://gofleye.com/}

\bibitem{uav_suvey_floreano_15}
Floreano, D., Wood, R.J.: Science, technology and the future of small
  autonomous drones.
\newblock Nature \textbf{521}, 460--466 (2015)

\bibitem{uav_product_flyability}
{Safe Drone for Inaccessible Places} (2014).
\newblock \urlprefix\url{https://www.flyability.com/}

\bibitem{display_interaction-gomes-chi16}
Gomes, A., Rubens, C., Braley, S., Vertegaal, R.: Bitdrones: {T}owards using
  {3D} nanocopter displays as interactive self-levitating programmable matter.
\newblock In: Proc. ACM Conference on Human Factors in Computing Systems (CHI),
  pp. 770--780 (2016)

\bibitem{uav_hri-platform_graether_12}
Graether, E., Mueller, F.: Joggobot: {A} flying robot as jogging companion.
\newblock In: Proc. CHI Extended Abstracts on Human Factors in Computing
  Systems, pp. 1063--1066 (2012)

\bibitem{hri_harper_ijsr09}
Harper, C., Virk, G.: Towards the development of international safety standards
  forhuman robot interaction.
\newblock International Journal of Social Robotics (IJSR) \textbf{2}(3),
  229--234 (2010).
\newblock \doi{10.1007/s12369-010-0051-1}.
\newblock \urlprefix\url{https://doi.org/10.1007/s12369-010-0051-1}

\bibitem{social-interaction_hedayati_hri18}
Hedayati, H., Walker, M., Szafir, D.: Improving collocated robot teleoperation
  with augmented reality.
\newblock In: Proc. ACM/IEEE International Conference on Human-Robot
  Interaction (HRI), pp. 78--86 (2018)

\bibitem{companion_uav-hepp-iros16}
Hepp, B., N{\"a}geli, T., Hilliges, O.: Omni-directional person tracking on a
  flying robot using occlusion-robust ultra-wideband signals.
\newblock In: Proc. IEEE/RSJ International Conference on Intelligent Robots and
  Systems (IROS), pp. 189--194 (2016)

\bibitem{uav_hri-human-tracking_higuchi_11}
Higuchi, K., Ishiguro, Y., Rekimoto, J.: Flying eyes: {F}ree-space content
  creation using autonomous aerial vehicles.
\newblock In: Proc. ACM Conference on Human Factors in Computing Systems (CHI),
  pp. 561--570 (2011)

\bibitem{companion_uav-hrabia-ros17}
Hrabia, C.E., Berger, M., Hessler, A., Wypler, S., Brehmer, J., Matern, S.,
  Albayrak, S.: An autonomous companion {UAV} for the {SpaceBot Cup Competition
  2015}.
\newblock In: Robot Operating System (ROS): The Complete Reference (Volume 2),
  pp. 345--385 (2017)

\bibitem{uav_hri-interface_huang_10}
Huang, A.S., Tellex, S., Bachrach, A., Kollar, T., Roy, D., Roy, N.: Natural
  language command of an autonomous micro-air vehicle.
\newblock In: Proc. IEEE/RSJ International Conference on Intelligent Robots and
  Systems (IROS), pp. 2663--2669 (2010)

\bibitem{hdi_huang_icra18}
Huang, C., Gao, F., Pan, J., Yang, Z., Qiu, W., Chen, P., Yang, X., Shen, S.,
  Cheng, K.T.: Act: An autonomous drone cinematography system for action
  scenes.
\newblock In: Proc. IEEE International Conference on Robotics and Automation
  (ICRA), pp. 7039--7046 (2018)

\bibitem{air_traffic_control-hutchins-ssrr07}
Hutchins, A.R., Cummings, M., Aubert, M.C., Uzumcu, S.C.: Toward the
  development of a low-altitude air traffic control paradigm for networks of
  small, autonomous unmanned aerial vehicles.
\newblock In: Proc. AIAA Infotech @ Aerospace, pp. 1--8 (2015)

\bibitem{hdi_jang_ral19}
Jang, J., Cho, K., Yang, G.H.: Design and experimental study of
  dragonfly-inspired flexible blade to improve safety of drones.
\newblock IEEE Robotics and Automation Letters (RA-L) \textbf{4}(4), 4200--4207
  (2019)

\bibitem{social-interaction_jensen_chi18}
Jensen, W., Hansen, S., Knoche, H.: Knowing you, seeing me: Investigating user
  preferences in drone-human acknowledgement.
\newblock In: Proc. ACM Conference on Human Factors in Computing Systems (CHI),
  pp. 1--12 (2018)

\bibitem{social-robot_jibo_times17}
{Jibo Robot - The 25 Best Inventions of 2017} (2017).
\newblock \urlprefix\url{https://time.com/5023212/best-inventions-of-2017/}

\bibitem{uav_hri-design_noise-issue-1}
Johnson, D., Papadopoulos, P., Watfa, N., Takala, J.: Exposure criteria:
  {O}occupational exposure levels.
\newblock In: B.~Goelzer, C.H. Hansen, G.A. Sehrndt (eds.) Occupational
  exposure to noise: {E}valuation, prevention and control, chap.~4, pp.
  79--102. Federal Institute for Occupational Safety and Health (2001)

\bibitem{social-interaction_jones_dis16}
Jones, B., Dillman, K., Tang, R., Tang, A., Sharlin, E., Oehlberg, L.,
  Neustaedter, C., Bateman, S.: Elevating communication, collaboration, and
  shared experiences in mobile video through drones.
\newblock In: Proc. ACM Conference on Designing Interactive Systems (DIS), pp.
  1123--1135 (2016)

\bibitem{social-robot_csaba_jais18}
Kert{\'e}sz, C., Turunen, M.: Exploratory analysis of sony {AIBO} users.
\newblock Journal of {AI} \& Society \textbf{34}(3), 625--638 (2019)

\bibitem{social-interaction_khan_chi19}
Khan, M.N.H., Neustaedter, C.: An exploratory study of the use of drones for
  assisting firefighters during emergency situations.
\newblock In: Proc. ACM Conference on Human Factors in Computing Systems (CHI),
  pp. 1--14 (2019)

\bibitem{hdi_khan_chi19}
Khan, M.N.H., Neustaedter, C.: An exploratory study of the use of drones for
  assisting firefighters during emergency situations.
\newblock In: Proc. ACM Conference on Human Factors in Computing Systems (CHI),
  pp. 1--14 (2019)

\bibitem{sound_interaction-kim-jpc17}
Kim, S., Chang, K.J., Park, D.C., Lee, S.M., Lee, S.K.: A systematic approach
  to engine sound design for enhancing sound character by active sound design.
\newblock SAE International Journal of Passenger Cars - Mechanical Systems
  \textbf{10}(3), 691--702 (2017)

\bibitem{social-interaction_kljun_chiplay15}
Kljun, M., Pucihar, K.{\v C}., Lochrie, M., Egglestone, P.: Streetgamez: A
  moving projector platform for projected street games.
\newblock In: Proc. Annual Symposium on Computer-Human Interaction in Play (CHI
  PLAY), pp. 589--594 (2015)

\bibitem{hdi_knierim_tei18}
Knierim, P., Kosch, T., Achberger, A., Funk, M.: Flyables: Exploring 3d
  interaction spaces for levitating tangibles.
\newblock In: Proc. International Conference on Tangible, Embedded, and
  Embodied Interaction (TEI), pp. 329--336 (2018)

\bibitem{hdi_knierim_chi18}
Knierim, P., Maurer, S., Wolf, K., Funky, M.: Quadcopter-projected in-situ
  navigation cues for improved location awareness.
\newblock In: Proc. ACM Conference on Human Factors in Computing Systems (CHI),
  pp. 1--6 (2018)

\bibitem{challenges-kornatowski-iros17}
Kornatowski, P.M., Mintchev, S., Floreano, D.: An origami-inspired cargo drone.
\newblock In: Proc. IEEE/RSJ International Conference on Intelligent Robots and
  Systems (IROS), pp. 6855--6862 (2017)

\bibitem{hdi_kuznetsov_ubicomp11}
Kuznetsov, S., Davis, G.N., Paulos, E., Gross, M.D., Cheung, J.C.: Red balloon,
  green balloon, sensors in the sky.
\newblock In: Proc. ACM International Joint Conference on Pervasive and
  Ubiquitous Computing (UbiComp), pp. 237--246 (2011)

\bibitem{uav-background_quadcopter_latscha_iros14}
Latscha, S., Kofron, M., Stroffolino, A., Davis, L., Merritt, G., Piccoli, M.,
  Yim, M.: Design of a hybrid exploration robot for air and land deployment
  {(H.E.R.A.L.D)} for urban search and rescue applications.
\newblock In: Proc. IEEE/RSJ International Conference on Intelligent Robots and
  Systems (IROS), pp. 1868--1873 (2014)

\bibitem{uav_hri-design_lee_17}
Lee, J.Y., Song, S.H., Shon, H.W., Choi, H.R., Yim, W.: Modeling and control of
  a saucer type {C}oand\u{a} effect {UAV}.
\newblock In: Proc. IEEE International Conference on Robotics and Automation
  (ICRA), pp. 2717--2722 (2017)

\bibitem{uav_hri-interface_lichtenstern_12}
Lichtenstern, M., Frassl, M., Perun, B., Angermann, M.: A prototyping
  environment for interaction between a human and a robotic multi-agent system.
\newblock In: Proc. ACM/IEEE International Conference on Human-Robot
  Interaction (HRI), pp. 185--186 (2012)

\bibitem{liew_dissertation_16}
Liew, C.F.: Towards human-robot interaction in flying robots: {A} user
  accompanying model and a sensing interface.
\newblock Ph.D. thesis, Department of Aeronautics and Astronautics, The
  University of Tokyo, Japan (2016)

\bibitem{uav-survey_liew_arxiv_17}
Liew, C.F., DeLatte, D., Takeishi, N., Yairi, T.: Recent developments in aerial
  robotics: {A} survey and prototypes overview.
\newblock ArXiv e-prints  (2017)

\bibitem{facial_expression-liew-jface12}
Liew, C.F., Yokoya, N., Yairi, T.: Control of unmanned aerial vehicle using
  facial expression.
\newblock In: Proc. Japanese Academy of Facial Studies, pp. 1--1 (2012)

\bibitem{uav_hri-human-tracking_lim_15}
Lim, H., Sinha, S.N.: Monocular localization of a moving person onboard a
  quadrotor {MAV}.
\newblock In: Proc. IEEE International Conference on Robotics and Automation
  (ICRA), pp. 2182--2189 (2015)

\bibitem{sound_interaction-lyon-jas00}
Lyon, R.H.: Product sound quality: From perception to design.
\newblock Sound and Vibration \textbf{37}(3), 18--23 (2003)

\bibitem{companion_uav-malliaraki-vimeo17}
Natural human-drone interaction (2017).
\newblock \urlprefix\url{https://emalliaraki.com/social-drones}

\bibitem{uav_hri-design_noise-issue-3}
Maue, J.: {Noise - European Agency for Safety and Health at Work} (2018).
\newblock \urlprefix\url{http://osha.europa.eu/en/topics/noise/}

\bibitem{hdi_pour_iros18}
MohaimenianPour, S., Vaughan, R.: Hands and faces, fast: Mono-camera user
  detection robust enough to directly control a uav in fligh.
\newblock In: Proc. IEEE/RSJ International Conference on Intelligent Robots and
  Systems (IROS), pp. 5224--5231 (2018)

\bibitem{uav_hri-interface_monajjemi_14}
Monajjemi, V.M., Pourmehr, S., Sadat, S.A., Zhan, F., Wawerla, J., Mori, G.,
  Vaughan, R.: Integrating multi-modal interfaces to command {UAV}s.
\newblock In: Proc. ACM/IEEE International Conference on Human-Robot
  Interaction (HRI), pp. 106--106 (2014)

\bibitem{uav_hri-interface_monajjemi_13}
Monajjemi, V.M., Wawerla, J., Vaughan, R., Mori, G.: {HRI} in the sky:
  {C}reating and commanding teams of {UAV}s with a vision-mediated gestural
  interface.
\newblock In: Proc. IEEE/RSJ International Conference on Intelligent Robots and
  Systems (IROS), pp. 617--623 (2013)

\bibitem{sound_interaction-pena-hri17}
Moore, D., Tennent, H., Martelaro, N., Ju, W.: Making noise intentional: {A}
  study of servo sound perception.
\newblock In: Proc. ACM/IEEE International Conference on Human-Robot
  Interaction (HRI), pp. 12--21 (2017)

\bibitem{uav_hri-safety_moskowitz_18}
Moskowitz, E.E., Siegel-Richman, Y.M., Hertner, G., Schroeppel, T.: Aerial
  drone misadventure: A novel case of trauma resulting in ocular globe rupture.
\newblock American Journal of Ophthalmology Case Reports \textbf{10}, 35--37
  (2018)

\bibitem{companion_uav-murphy-ar11}
Murphy, R., Shell, D., Guerin, A., Duncan, B., Fine, B., Pratt, K., Zourntos,
  T.: A midsummer night’s dream (with flying robots).
\newblock Autonomous Robots \textbf{30}(2), 143--156 (2011)

\bibitem{hdi_nagi_hri14}
Nagi, J., Giusti, A., Caro, G.A.D., Gambardella, L.M.: Human control of uavs
  using face pose estimates and hand gestures.
\newblock In: Proc. ACM/IEEE International Conference on Human-Robot
  Interaction (HRI), pp. 1--2 (2014)

\bibitem{uav_hri-human-tracking_naseer_13}
Naseer, T., Sturm, J., Cremers, D.: {FollowMe}: {P}erson following and gesture
  recognition with a quadrocopter.
\newblock In: Proc. IEEE/RSJ International Conference on Intelligent Robots and
  Systems (IROS), pp. 624--630 (2013)

\bibitem{uav_hri-social-study_ng_11}
Ng, W.S., Sharlin, E.: Collocated interaction with flying robots.
\newblock In: Proc. IEEE International Symposium on Robot and Human Interactive
  Communication (RO-MAN), pp. 143--149 (2011)

\bibitem{uav_hri-platform_nitta_14}
Nitta, K., Higuchi, K., Rekimoto, J.: {HoverBall}: {A}ugmented sports with a
  flying ball.
\newblock In: Proc. Augmented Human International Conference, pp. 13:1--13:4
  (2014)

\bibitem{sound_interaction-norman-book13}
Norman, D.A.: The Design of Everyday Things: Revised and Expanded Edition.
\newblock Basic Books (2013)

\bibitem{hdi_nowacka_ubicomp15}
Nowacka, D., Hammerla, N.Y., Elsden, C., Pl{\"o}tz, T., Kirk, D.: Diri - the
  actuated helium balloon: A study of autonomous behaviour in interfaces.
\newblock In: Proc. ACM International Joint Conference on Pervasive and
  Ubiquitous Computing (UbiComp), pp. 349--360 (2015)

\bibitem{display_interaction-nozaki-chi14}
Nozaki, H.: Flying display: {A} movable display pairing projector and screen in
  the air.
\newblock In: Proc. ACM Conference on Human Factors in Computing Systems (CHI),
  pp. 909--914 (2014)

\bibitem{thermal_interaction-pena-hri18}
Pachamango, D.P., Tanaka, F.: Touch to feel me: {D}esigning a robot for
  thermo-emotional communication.
\newblock In: Proc. ACM/IEEE International Conference on Human-Robot
  Interaction (HRI), pp. 207--208 (2018)

\bibitem{uav-background_quadcopter_papachristos_iros15}
Papachristos, C., Tzoumanikas, D., Tzes, A.: Aerial robotic tracking of a
  generalized mobile target employing visual and spatio-temporal dynamic
  subject perception.
\newblock In: Proc. IEEE/RSJ International Conference on Intelligent Robots and
  Systems (IROS), pp. 4319--4324 (2015)

\bibitem{thermal_interaction-park-ro14}
Park, E., Lee, J.: I am a warm robot: {T}he effects of temperature in physical
  human–robot interaction.
\newblock Robotica \textbf{32}(1), 133--142 (2014)

\bibitem{uav_product_ar-drone}
{Parrot AR.Drone} (2015).
\newblock \urlprefix\url{https://en.wikipedia.org/wiki/Parrot_AR.Drone}

\bibitem{uav_product_ar-drone-2}
{Parrot AR.Drone 2.0} (2015).
\newblock \urlprefix\url{http://www.parrot.com/products/ardrone-2/}

\bibitem{gesture_interaction-peshkova-pc17}
Peshkova, E., Hitz, M., Kaufmann, B.: Natural interaction techniques for an
  unmanned aerial vehicle system.
\newblock IEEE Pervasive Computing \textbf{16}(1), 34--42 (2017)

\bibitem{uav_hri-human-tracking_pestana_13}
Pestana, J., Sanchez-Lopez, J.L., Campoy, P., Saripalli, S.: Vision based
  {GPS}-denied object tracking and following for unmanned aerial vehicles.
\newblock In: Proc. IEEE International Symposium on Safety, Security, and
  Rescue Robotics (SSRR), pp. 1--6 (2013)

\bibitem{gesture_interaction-pfeil-iui13}
Pfeil, K.P., Koh, S.L., Jr., J.J.L.: Exploring {3D} gesture metaphors for
  interaction with unmanned aerial vehicles.
\newblock In: Proc. ACM International Conference on Intelligent User Interfaces
  (IUI), pp. 257--266 (2013)

\bibitem{public_issue-pytlikZillig-tsm18}
PytlikZillig, L.M., Duncan, B., Elbaum, S., Detweiler, C.: A drone by any other
  name.
\newblock IEEE Technology and Society Magazine \textbf{37}(1), 80--91 (2018)

\bibitem{sound_interaction-read-dissertation14}
Read, R.: A study of non-linguistic utterances for social human-robot
  interaction.
\newblock Ph.D. thesis, Faculty of Science and Environment, Plymouth
  University, UK (2014)

\bibitem{safety_concerns-mintchev-ral17}
S.~Mintchev, S.d.R., Floreano, D.: Insect-inspired mechanical resilience for
  multicopters.
\newblock IEEE Robotics and Automation Letters \textbf{2}(3), 1248--1255 (2017)

\bibitem{hdi_salaan_ral19}
Salaan, C.J., Tadakuma, K., Okada, Y., Sakai, Y., Ohno, K., Tadokoro, S.:
  Development and experimental validation of aerial vehicle with passive
  rotating shell on each rotor.
\newblock IEEE Robotics and Automation Letters (RA-L) \textbf{4}(3), 2568--2575
  (2019)

\bibitem{hdi_scheible_perdis13}
Scheible, J., Hoth, A., Saal, J., Su, H.: Displaydrone: A flying robot based
  interactive display.
\newblock In: Proc. ACM International Symposium on Pervasive Displays (PerDis),
  pp. 49--54 (2013)

\bibitem{display_interaction-schneegass-chi14}
Schneegass, S., Alt, F., Scheible, J., Schmidt, A., Su, H.: Midair displays:
  {E}xploring the concept of free-floating public displays.
\newblock In: Proc. ACM Conference on Human Factors in Computing Systems (CHI),
  pp. 2035--2040 (2014)

\bibitem{uav_hri-social-study_sharma_13}
Sharma, M., Hildebrandt, D., Newman, G., Young, J.E., Eskicioglu, R.:
  Communicating affect via flight path: {E}xploring use of the laban effort
  system for designing affective locomotion paths.
\newblock In: Proc. ACM/IEEE International Conference on Human-Robot
  Interaction (HRI), pp. 293--300 (2013)

\bibitem{uav_hri-design_song_17}
Song, S., Yamada, S.: Expressing emotions through color, sound, and vibration
  with an appearance-constrained social robot.
\newblock In: Proc. ACM/IEEE International Conference on Human-Robot
  Interaction (HRI), pp. 2--11 (2017)

\bibitem{hdi_song_iros18}
Song, S.H., Shon, H.W., Yeon, G.Y., Choi, H.R.: Design and implementation of
  cloud-like soft drone s-cloud.
\newblock In: Proc. IEEE/RSJ International Conference on Intelligent Robots and
  Systems (IROS), pp. 1--9 (2018)

\bibitem{hdi_soto_assets17}
Soto, M.A., Funk, M., Hoppe, M., Boldt, R., Wolf, K., Henze, N.:
  Dronenavigator: Using leashed and free-floating quadcopters to navigate
  visually impaired travelers.
\newblock In: Proc. International ACM SIGACCESS Conference on Computers and
  Accessibility (ASSETS), pp. 300--304 (2017)

\bibitem{uav_hri-platform_flying-lampshade-1}
{SPARKED: A Live Interaction Between Humans and Quadcopters} (2016).
\newblock \urlprefix\url{https://www.youtube.com/watch?v=6C8OJsHfmpI}

\bibitem{uav_hri-platform_flying-lampshade-2}
{SPARKED: Behind the Technology} (2016).
\newblock \urlprefix\url{https://www.youtube.com/watch?v=7YqUocVcyrE}

\bibitem{uav_hri-interface_sun_17}
Sun, T., Nie, S., Yeung, D.Y., Shen, S.: Gesture-based piloting of an aerial
  robot using monocular vision.
\newblock In: Proc. IEEE International Conference on Robotics and Automation
  (ICRA), pp. 5913--5920 (2017)

\bibitem{uav_hri-social-study_szafir_14}
Szafir, D., Mutlu, B., Fong, T.: Communication of intent in assistive free
  flyers.
\newblock In: Proc. ACM/IEEE International Conference on Human-Robot
  Interaction (HRI), pp. 358--365 (2014)

\bibitem{light_interaction-szafir-hri15}
Szafir, D., Mutlu, B., Fong, T.: Communicating directionality in flying robots.
\newblock In: Proc. ACM/IEEE International Conference on Human-Robot
  Interaction (HRI), pp. 19--26 (2015)

\bibitem{online_application_tagspaces}
{TagSpaces - Your Hackable File Organizer} (2017).
\newblock \urlprefix\url{http://thedroneracingleague.com/}

\bibitem{social-robot_tallec_sbh11}
Tallec, M.L., Saint-Aim{\'e}, S., Jost, C., Villaneau, J., Antoine, J.Y.,
  Letellier-Zarshenas, S., Le-P{\'e}v{\'e}dic, B., Duhaut, D.: From speech to
  emotional interaction: {EmotiRob} project.
\newblock Human-Robot Personal Relationships pp. 57--64 (2011)

\bibitem{uav_hri_safety-drone_accident}
{Toddler's Eyeball Sliced in Half by Drone Propeller} (2015).
\newblock
  \urlprefix\url{https://www.bbc.com/news/uk-england-hereford-worcester-34936739/}

\bibitem{public_issue-un-ocha14}
{United Nations OCHA}: Unmanned aerial vehicles in humanitarian response.
\newblock Tech. rep., OCHA Policy and Studies Series (2014)

\bibitem{handbook_of_uav}
Valavanis, K.P., Vachtsevanos, G.J.: Handbook of Unmanned Aerial Vehicles.
\newblock Springer Publishing Company, Incorporated (2014)

\bibitem{uav_product_snap}
{Snap drone} (2016).
\newblock \urlprefix\url{https://vantagerobotics.com/}

\bibitem{social-interaction_vink_youtube14}
Vink, L., Cauchard, J., Landay, J.A.: {Autonomous Wandering Interface (AWI) -
  Concept Video} (2014).
\newblock \urlprefix\url{https://www.youtube.com/watch?v=cqU_hR2_ILU}

\bibitem{social-robot_wada_tro07}
Wada, K., Shibata, T.: Living with seal robots—its sociopsychological and
  physiological influences on the elderly at a care house.
\newblock IEEE Transactions on Robotics \textbf{23}(5), 972--980 (2007)

\bibitem{hdi_walker_hri18}
Walker, M., Hedayati, H., Lee, J., Szafir, D.: Communicating robot motion
  intent with augmented reality.
\newblock In: Proc. ACM/IEEE International Conference on Human-Robot
  Interaction (HRI), pp. 316--324 (2018)

\bibitem{social-interaction_wojciechowska_imwut19}
Wojciechowska, A., Frey, J., Mandelblum, E., Amichai-Hamburger, Y., Cauchard,
  J.R.: Designing drones: Factors and characteristics influencing the
  perception of flying robots.
\newblock Proc. ACM Interactive, Mobile, Wearable and Ubiquitous Technologies
  (IMWUT) \textbf{3}(3), 1--19 (2019)

\bibitem{social-interaction_wojciechowska_hri19}
Wojciechowska, A., Frey, J., Sass, S., Shafir, R., Cauchard, J.R.: Collocated
  human-drone interaction: Methodology and approach strategy.
\newblock In: Proc. ACM/IEEE International Conference on Human-Robot
  Interaction (HRI), pp. 172--181 (2019)

\bibitem{hdi_yamada_chi19}
Yamada, W., Manabe, H., Ikeda, D.: {ZeRONE}: Safety drone with blade-free
  propulsion.
\newblock In: Proc. ACM Conference on Human Factors in Computing Systems (CHI),
  pp. 1--8 (2019)

\bibitem{uav_hri-human-tracking_yao_17}
Yao, N., Anaya, E., andd Sungjin~Cho, Q.T., Zheng, H., Zhang, F.: Monocular
  vision-based human following on miniature robotic blimp.
\newblock In: Proc. IEEE International Conference on Robotics and Automation
  (ICRA), pp. 3244--3249 (2017)

\bibitem{social-interaction_yeh_hai17}
Yeh, A., Ratsamee, P., Kiyokawa, K., Uranishi, Y., Mashita, T., Takemura, H.,
  Fjeld, M., Obaid, M.: Exploring proxemics for human-drone interaction.
\newblock In: Proc. International Conference on Human Agent Interaction (HAI),
  pp. 81--88 (2017)

\bibitem{hri_yohanan_ijsr12}
Yohanan, S., MacLean, K.E.: The role of affective touch in human-robot
  interaction: Human intent and expectations in touching the haptic creature.
\newblock International Journal of Social Robotics (IJSR) \textbf{4}(2),
  163--180 (2012).
\newblock \doi{10.1007/s12369-011-0126-7}.
\newblock \urlprefix\url{https://doi.org/10.1007/s12369-011-0126-7}

\bibitem{uav_product_hover-camera}
{Hover Camera} (2016).
\newblock \urlprefix\url{http://gethover.com/}

\bibitem{hdi_zhou_icra18}
Zhou, X., Liu, S., Pavlakos, G., Kumar, V., Daniilidis, K.: Human motion
  capture using a drone.
\newblock In: Proc. IEEE International Conference on Robotics and Automation
  (ICRA), pp. 2027--2033 (2018)

\end{thebibliography}

%\vspace{4mm}
\noindent
\textbf{Chun Fui Liew} received his B.Eng. degree from the Nanyang Technological University, Singapore and M.Sc. degree from the National University of Singapore. He also received his M.Eng. and Ph.D. degrees in Aeronautics and Astronautics Engineering from the University of Tokyo, Japan. He has worked as a drone project leader and researcher with the Hongo Aerospace Inc., Japan. He is currently a full-time post-doctoral researcher in the University of Tokyo. His research interests include aerial robotics, pattern recognition, machine learning, and computer vision.

\vspace{2mm}
\noindent
\textbf{Takehisa Yairi} received his M.Eng. and Ph.D. degrees from the University of Tokyo, Japan in 1996 and 1999 respectively. He is currently a full-time Associate Professor with the Graduate School of Engineering in the University of Tokyo. His research interests include data mining, machine learning, mobile and space robotics. He is a member of The Japanese Society for Artificial Intelligence (JSAI) and The Robotics Society of Japan (RSJ).

\end{document}